# An Efficient Method for Face Recognition System In Various Assorted Conditions

V.Karthikeyan      K.Vijayalakshmi      P.Jeyakumar

**Abstract**-In the beginning stage, face verification is done using easy method of geometric algorithm models, but the verification route has now developed into a scientific progress of complicated geometric representation and identical procedure. In recent years the technologies have boosted face recognition system into the healthy focus. Researcher's currently undergoing strong research on finding face recognition system for wider area information taken under hysterical elucidation dissimilarity. The proposed face recognition system consists of a narrative exposition-indiscreet preprocessing method, a hybrid Fourier-based facial feature extraction and a score fusion scheme. We have verified the face recognition in different lightening conditions (day or night) and at different locations (indoor or outdoor). Preprocessing, Image detection, Feature- extraction and Face recognition are the methods used for face verification system. This paper focuses mainly on the issue of toughness to lighting variations. The proposed system has obtained an average of 88.1% verification rate on Two-Dimensional images under different lightening conditions.

## I.INTRODUCTION

In the past decades, many appearance-based methods have been proposed to handle this problem, and new theoretical insights as well as good recognition results have been normalization, feature extraction, and subspace representation, each stage increases resistance to illumination variations and makes the information needed for recognition more manifest. This method achieves very significant improvements, than the other method of verification rate is 88.1% at 0.1% false acceptance rate. Several aspects of the relationship between image normalization and feature sets, Robust feature sets and feature comparison strategies, Fusion of multiple feature sets framework is the  will be verified.

reported. In the proposed the verification of the face in different climatic conditions this paper focuses mainly on the issue of robustness to lighting variations. Traditional approaches for dealing with this issue can be broadly classified into three categories: appearance-based, normalization-based, and feature-based methods. In direct appearance-based approaches, training examples are collected under different lighting conditions and directly (i.e., without undergoing any lighting preprocessing) used to learn a global model of the possible illumination variations but it requires a large number of training images and an expressive feature set, otherwise it is essential to include a good preprocessor to reduce illumination variations. Normalization based approaches seek to reduce the image to a more "canonical" form in which the illumination variations are suppressed. Histogram equalization is one simple example of this method. These methods are quite effective but their ability to handle spatially non uniform variations remains limited. The third approach extracts illumination-insensitive feature sets directly from the given image. These feature sets range from geometrical features to image derivative features such as edge maps, local binary patterns (LBP), Gabor wavelets, and local autocorrelation filters. Although such features offer a great improvement on raw gray values, their resistance to the complex illumination variations that occur in real-world face images is still quite limited. The integrative framework is proposed, that combines the strengths of all three of the above approaches. The overall process can be viewed as a pipeline consistingofimage

---

[1] **Prof.V.Karthikeyan**  Department of Electronics and Communication Engineering SVS College of Engineering and Technology, Coimbatore-642109  E-mail:karthick77keyan@gmail.com

**Prof.K.Vijayalakshmi**, Assistant Professor, Department of Electrical and Electronics Engineering, SSK College of Engineering & Technology, Coimbatore-641105 E-Mail: vijik810@gmail.com

**P.Jeyakumar** Department of Electronics and Communication Engineering SVS College of Engineering and Technology, Coimbatore-642109

## II. LITERATURE REVIEW

There are two predominant approaches to the face recognition problem: Geometric (feature based) and photometric (view based).Many different algorithms were developed PCA commonly referred to as the use of Eigen faces, is the technique With PCA, the probe and gallery images must be the same size and must first be normalized to line up the eyes and mouth of the subjects within the image. The PCA approach is then used to reduce the dimension of the data by means of data compression basics and reveals the most effective low dimensional structure of facial patterns. This reduction in dimensions removes information that is not useful and precisely decomposes the face structure into orthogonal components known as Eigen faces. Each face image may be represented as the weighted sum of the Eigen faces, which are stored in a 1D array. A probe image is compared against a gallery image by measuring the distance between their respective feature vectors. The PCA approach typically requires the full frontal face to be presented each time; otherwise the image results in poor performance. The primary advantage of this technique is that it can reduce the data needed to identify the individual to $1/1000^{th}$ of the data

Presented.

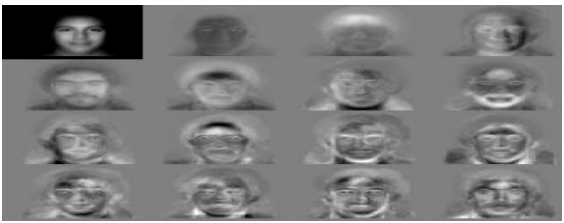

Fig: 1 Standard Eigen faces

LDA is a statistical approach for classifying samples of unknown classes based on training samples with classes. This technique aims to maximize between-class variance and minimize within –class variance. In fig 2 where each block represents a class, there are large variances between classes, but little variance with in classes. When dealing with high dimensional face data, this technique faces the small sample size problem that arises where there are a small number of available training samples compared to the dimensionality of the sample space.

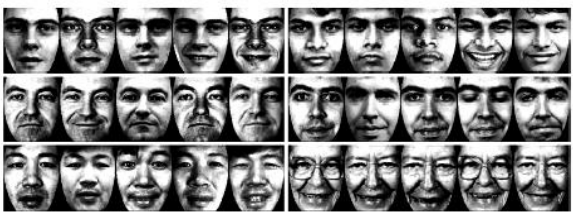

Fig: 2 Examples of six classes using LDA

EGBM relies on the concept that real face images have many non linear characteristics that are not addressed by the other linear methods, which includes variation in illumination, pose and expression. A Gabor wavelet transform creates a dynamic link architecture that projects the face onto an elastic grid. The Gabor jet is a node on the elastic grid notated by circles on the image below which describes the image behavior around a given pixel. It is the result of a convolution of the image with a Gabor filter, which is used to detect shapes and to extract features using image processing. Recognition is based on the similarity of the Gabor filter response at each Gabor node. The difficulty with this method is the requirement of accurate land mark localization which can sometimes be achieved by combining PCA and LDA methods.

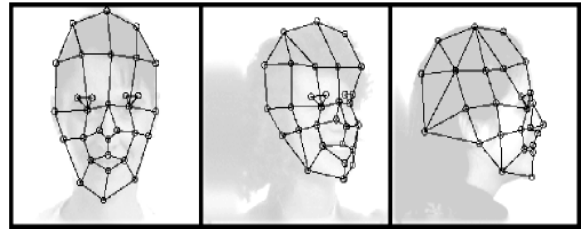

Fig: 3 Elastic bunch map graphing\

In image processing and computer vision , anisotropic diffusion, also called Perona–Malik diffusion, is a technique aiming at reducing image noise without removing significant parts of the image content, typically edges, lines or other details that are important for the interpretation of the image. Anisotropic diffusion resembles the process that creates a scale-space, where an image generates a parameterized family of successively more and more blurred images based on a diffusion process. Each of the resulting images in this family is given as a convolution $c$ between the image and a 2D isotropic Gaussian filter, where the width of the filter increases with the parameter. This diffusion process is a linear and space invariant transformation of the original image. Anisotropic diffusion is a generalization of this diffusion process: it produces a family of parameterized images, but each resulting image is a combination between the original image and a filter that depends on the local content of the original image. As a consequence, anisotropic diffusion is a non-linear and space-variant transformation of the original image. An image gradient is a directional change in the intensity or color in an image. Image gradients may be used to extract information from images. Mathematically, the gradient of a two-variable function (here the image intensity function) is at each image point a 2D vector with the components given by the derivatives in the horizontal and vertical directions. At each image point, the gradient vector points in the direction of largest possible intensity increase, and the length of the gradient vector corresponds to the rate of change in that direction.Since the intensity function of a digital image is only known at discrete points, derivatives of this function cannot be defined unless we assume that there is an underlying continuous intensity function which has been sampled at the image points. With some additional assumptions, the derivative of the continuous intensity function can be computed as a function on the sampled intensity function, i.e., the digital image. It turns out that the derivatives at any particular point are functions of the intensity values at virtually all image points. However, approximations of these derivative functions can be defined at lesser or larger degrees of accuracy. The sobel operator represents a rather inaccurate approximation of the image gradient, but is still of sufficient quality to be of practical use in many applications. More precisely, it uses intensity values only in a 3×3 region around each image point to approximate the corresponding image gradient, and it uses only integer values for the coefficients which weight the image intensities to produce the gradient approximation.

## III. PROPOSED SYSTEM

We can make the following assumptions: 1) most of the intrinsic factor is in the high spatial frequency domain, and 2) most of the extrinsic factor is in the low spatial frequency domain. Considering the first assumption, one might use a high-pass filter to extract the intrinsic factor, but it has been proved that this kind of filter is not robust to illumination variations as shown. In addition, a high-pass filtering operation may have a risk of removing some of the useful intrinsic factor. Hence, we propose an alternative approach, namely, employing a gradient operation. The approximation comes from the assumptions that both the surface normal direction (shape) and the light source direction vary slowly across the image whereas the surface texture varies fast. The scaling factor is the extrinsic factor of our imaging model. The Retinex method and SQI method used the smoothed images as the estimation of this extrinsic factor. We also use the same approach to estimate the extrinsic part where is K a smoothing kernel and denotes the convolution. To overcome the illumination sensitivity, we normalized the gradient map with the following equation:

$$N = \frac{\nabla \chi}{\hat{W}} \approx \frac{(\nabla \rho) W}{\hat{W}} \approx \nabla \rho$$

After the normalization, the texture information in the normalized image N= {$N_x$, $N_y$} is still not apparent enough. In addition, the division operation may intensify unexpected noise terms. To recover the rich texture and remove the noise at the same time, we integrate the normalized gradients $N_x$ and $N_y$ with the anisotropic diffusion method which we explain in the following, and finally acquire the reconstructed image $X_r$

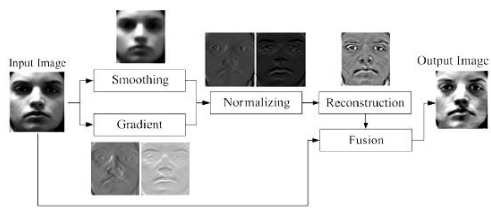

Fig: 4 Structure of the integral normalized gradient image

## IV. KERNEL PRINCIPLE COMPONENT ANALYSIS

KPCA encodes the pattern information based on second order dependencies, i.e., pixel wise covariance among the pixels, and are insensitive to the dependencies of multiple (more than two) pixels in the patterns. Since the eigenvectors in PCA are the orthonormal bases, the principal components are uncorrelated. In other words, the coefficients for one of the axes cannot be linearly represented from the coefficients of the other axes. Higher order dependencies in an image include nonlinear relations among the pixel intensity values, such as the relationships among three or more pixels in an edge or a curve, which can capture important information for recognition. Explicitly mapping the vectors in input space into higher dimensional space is computationally intensive. Using the kernel trick one can compute the higher order statistics using only dot products of the input patterns. Kernel PCA has been applied to face recognition applications and is observed to be able to extract nonlinear features. The advantage of using KPCA over other nonlinear feature extraction algorithms can be significant computationally. KPCA does not require solving a nonlinear optimization problem which is expensive computationally and the validity of the solution as optimal is typically a concern. KPCA only requires the solution of an eigenvalue problem. This reduces to using linear algebra to perform PCA in an arbitrarily large, possibly infinite dimensional, feature space. The kernel "trick" greatly simplifies calculations in this case. An additional advantage of KPCA is that the number of components does not have to be specified in advance. KPCA is a useful generalization that can be applied to these domains where nonlinear features require a nonlinear feature extraction tool. We plan to use the KPCA algorithm on real earth science data such as the sea surface temperature (SST) or normalized difference vegetation index (NDVI). The resulting information from KPCA can be correlated with signals such as the Southern Oscillation Index (SOI) for determining relationships with the El Nino phenomenon. KPCA can be used to discover nonlinear correlations in data that may otherwise not be found using standard PCA. The information generated about a data set using KPCA captures nonlinear features of the data. These features correlated with known spatial-temporal signals can discover nonlinear relationships. KPCA offers improved analysis of datasets that have nonlinear structure.

## V. LOG-LIKELY HOOD RATIO FOR SCORE FUSION

We interpret the set of scores as a feature vector from which we perform the classification task. We have a set of scores $S_1 \ldots S_n$ computed by classifiers. Now the problem is to decide whether the query-target pair is from the same person or not based upon these scores. We can cast this problem as the following hypothesis testing:

$H_0 : S_1 \ldots S_n \sim p(s_{1,\ldots,} sn|diff)$
$H_1 : S_1 \ldots S_n \sim p(s_{1,\ldots,} sn|same)$

Where p ($s_{1,\ldots,}$ sn|diff) is the distribution of the scores when the query and target are from different persons, and p ($s_{1,\ldots,}$ sn|same) is the distribution of the scores when the query and target are from the same person.

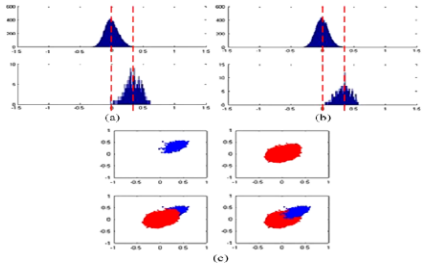

Fig: 5 Score distribution of two classifiers

The figure gives example of such distributions and provides the intuition behind the benefits of using multiple scores generated by multiple classifiers. Suppose we have two classifiers and they produce two scores $s_1$ and $s_2$ and Figure shows **p ($s_1$ |diff)** and **p ($s_1$ |same)** distributions of a single score. The region between the two vertical lines is where the two distributions overlap. Intuitively speaking, a classification error can occur when the score falls within this overlapped region, and the smaller this overlapped region, the smaller the probability of classification error. Likewise, Fig. 9(b) shows **p ($s_2$ |diff) and p ($s_2$ |same)** and. Fig(c) shows how the pair of the two scores ($s_1, s_2$) is distributed. The upper left of Fig(c) is the scatter. Plot ($s_1, s_2$) of when the query and target are from the same person and the upper right of Fig. 9(c) is the scatter plot of when the query and target are from different persons. The bottom of Fig(c) shows how the two scatter plots overlap. Compared with Fig. (a) and (b), we can see that the probability of overlap can be reduced by jointly considering the two scores $s_1$ and $s_2$ and which suggests that hypothesis testing based upon the two scores $s_1$ and $s_2$ is better than hypothesis testing based upon a single score $s_1$ $s_2$ .If we know the two densities **p ($s_1, \ldots, s_n$|diff)** and **p ($s_1, \ldots, s_n$|same)**, the log-likelihood ratio test achieves the highest verification rate for a given false accept rate $H_0: S_1 \ldots S_n \sim p(s_1,\ldots, s_n|diff)$

$$\text{Log} \frac{P(s_1,\ldots, s_n|same)}{P(s_1,\ldots, s_n|diff)} > < 0$$

However, the true densities p ($s_1, \ldots,$ $s_n$|diff) and p ($s_1, \ldots,$ $s_n$|same) are unknown, so we need to estimate these densities observing scores computed from query-target pairs in the training data. One way to estimate these densities is to use nonparametric density estimation. In this work, we use parametric density estimation in order to avoid over-fitting and reduce computational complexity. In particular, we model the distribution of $S_i$ given $H_0$ as a Gaussian random variable with mean $m_{diff,i}$ and variance $\sigma2_{diff,i}$, and model $\{S_i\}_{i=1}^n$ given $H_0$ as independent Gaussian random variables with density is given by

$$P(s_1,\ldots, s_n|diff) = \prod N(S_i; m_{diff,i}, \sigma2_{diff,i})$$

Where $N(S_i; m_{diff,i}, \sigma2) = (1/\sqrt{2\pi\sigma^2})\exp\{-((x-m)^2/2\sigma^2)\}$ is the Gaussian density function. The parameters $m_{diff, i}, \sigma2_{diff, i}$ are estimated from the scores of the $i^{th}$ classifier corresponding to no match query-target pairs in the training database. Similarly, we approximate the density of $\{S_i\}_{i=1}^n$ given $H_1$ by $\prod N(S_i; m_{same,i}, \sigma2_{same,i})$, and the parameters and $m_{diff, i}, \sigma2_{diff, i}$ and $m_{same, i}, \sigma2_{same, i}$ are computed from the scores of the classifier corresponding to match query-target pairs in the training database. Now we define the fused score to be the log-likelihood ratio, which is given by

$$S = \log \frac{N(S_i; m_{same,i}, \sigma^2_{same,i})}{N(S_i; m_{diff, i}, \sigma^2_{diff, i})}$$

VI. CONCLUSION

In the proposed approach we have through a whole face recognition system**.** The Retinex method and SQI method are used to smoothed images as the estimation of the extrinsic factor. KPCA offers improved analysis of datasets that have nonlinear structure. It is also useful for generalization that can be applied to the domains where nonlinear features are required. We also plan to use the KPCA algorithm on real earth science data. In this proposed work, we used parametric density estimation in order to avoid over-fitting and reduce the computational complexity with preprocessing, feature extraction and classifier and score fusion for uncontrolled illumination variations which will yield increased accuracy and efficiency. The overall process can be viewed as a pipeline consisting of image normalization, feature extraction, and subspace representation, each stage increases resistance to illumination variations and makes the information needed for recognition more manifest. This method achieves very significant improvements, than the other method of verification rate is 88.1% at 0.1% false acceptance rate. The integrative framework is proposed to verify the face in various lightening conditions.

VII. EXPERIMENTAL RESULTS"

The proposed work narrates the frame work of the experimental result as we take the gray scale image (0,255) as the original image. Fig 6 shows the Test image which is to be preprocessed. Fig 7 illustrates the preprocessed image 1 Fig.8 shows the image as we stored in the data base. Fig .9 demonstrates the final preprocessed image3.

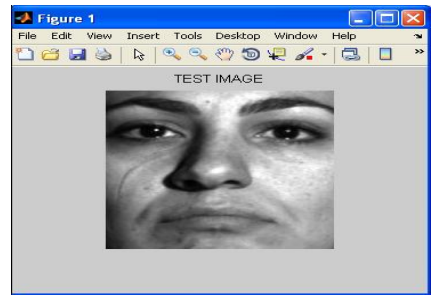

Fig. 6 Test image

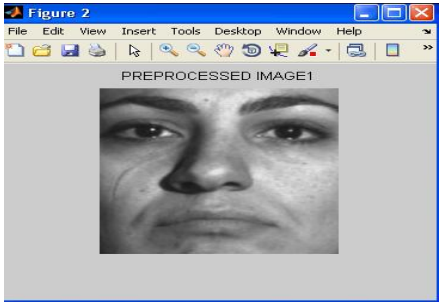

Fig .7 preprocessed image

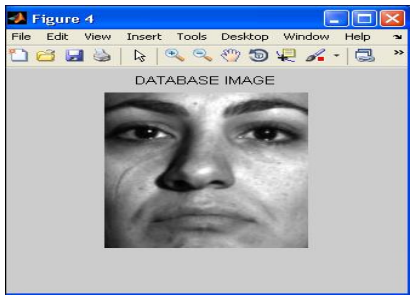

Fig. 8 Database image

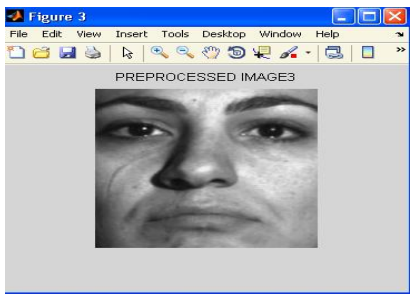

Fig. 9 Final preprocessed Image

## VIII. REFERENCES"

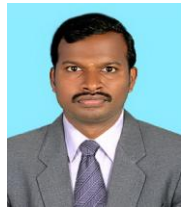

**Prof.V.Karthikeyan** has received his Bachelor's Degree in Electronics and Communication Engineering from PGP college of Engineering and technology in 2003 Namakkal, India. He received Masters Degree in Applied Electronics from KSR college of Technology, Erode in 2006. He is currently working as Assistant Professor in SVS College of Engineering and Technology, Coimbatore. She has about 7 years of teaching Experience.

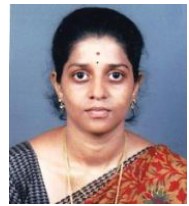

**Prof.K.Vijayalakshmi** has completed her Bachelor's Degree in Electrical & Electronics Engineering from Sri Ramakrishna Engineering College, Coimbatore, India. She finished her Masters Degree in Power Systems Engineering from Anna University of Technology, Coimbatore. She is currently working as Assistant Professor in Sri Krishna college of Engineering and Technology, Coimbatore. She has about 5 years of teaching Experience.